\documentclass[letterpaper]{article} 
\usepackage{aaai25}  
\usepackage{times}  
\usepackage{helvet}  
\usepackage{courier}  
\usepackage[hyphens]{url}  
\usepackage{graphicx} 
\usepackage{natbib}  
\usepackage{caption} 
\usepackage{soul}
\usepackage{graphicx}
\usepackage{caption}
\usepackage{lipsum}
\usepackage{amsthm}
\usepackage{amssymb}
\usepackage{algorithm}
\usepackage{footmisc} 
\usepackage{algorithmic}
\usepackage[algo2e, boxed, ruled]{algorithm2e}
\usepackage{booktabs}
\usepackage{amsmath}
\usepackage{placeins}
\usepackage{cleveref}
\usepackage{newfloat}
\usepackage{listings}
\DeclareCaptionStyle{ruled}{labelfont=normalfont,labelsep=colon,strut=off} 
\floatstyle{ruled}
\newfloat{listing}{tb}{lst}{}
\floatname{listing}{Listing}
\title{CustomCrafter: Customized Video Generation with Preserving Motion and Concept Composition Abilities}

\newcommand*\samethanks[1][\value{footnote}]{\footnotemark[#1]}

\author {
    Tao Wu \textsuperscript{\rm 1},
    Yong Zhang \textsuperscript{\rm 2}\thanks{Corresponding author.},
    Xintao Wang \textsuperscript{\rm 2,4},
    Xianpan Zhou \textsuperscript{\rm 3},
    Guangcong Zheng \textsuperscript{\rm 1},
    Zhongang Qi \textsuperscript{\rm 4},
    Ying Shan\textsuperscript{\rm 2,4},
    Xi Li \textsuperscript{\rm 1}\samethanks[1]
}
\affiliations {
    \textsuperscript{\rm 1}College of Computer Science and Technology, Zhejiang University \textsuperscript{\rm 2}Tencent AI Lab\\
    \textsuperscript{\rm 3}Polytechnic Institute, Zhejiang University
    \textsuperscript{\rm 4}ARC Lab, Tencent PCG \\
    taowucs@zju.edu.cn, \{norriszhang,xintaowang\}@tencent.com, \{zhouxianpan,guangcongzheng\}@zju.edu.cn, \{zhongangqi, yingsshan\}@tencent.com, xilizju@zju.edu.cn
}

\usepackage{bibentry}

\begin{document}
\maketitle




\begin{abstract}
    Customized video generation aims to generate high-quality videos guided by text prompts and subject's reference images.
    However, since it is only trained on static images, the fine-tuning process of subject learning disrupts abilities of video diffusion models (VDMs) to combine concepts and generate motions. 
    To restore these abilities, some methods use additional video similar to the prompt to fine-tune or guide the model. 
    This requires frequent changes of guiding videos and even re-tuning of the model when generating different motions, which is very inconvenient for users.
    In this paper, we propose CustomCrafter, a novel framework that preserves the model's motion generation and conceptual combination abilities without additional video and fine-tuning to recovery.
    For preserving conceptual combination ability, we design a plug-and-play module to update few parameters in VDMs, enhancing the model's ability to capture the appearance details and the ability of concept combinations for new subjects.
    For motion generation, we observed that VDMs tend to restore the motion of video in the early stage of denoising, while focusing on the recovery of subject details in the later stage. 
    Therefore, we propose Dynamic Weighted Video Sampling Strategy. 
    Using the pluggability of our subject learning modules, we reduce the impact of this module on motion generation in the early stage of denoising, preserving the ability to generate motion of VDMs. 
    In the later stage of denoising, we restore this module to repair the appearance details of the specified subject, thereby ensuring the fidelity of the subject's appearance.
    Experimental results show that our method has a significant improvement compared to previous methods.
    Code is available at \url{https://github.com/WuTao-CS/CustomCrafter}
\end{abstract}

\section{Introduction}
\begin{figure}[tb]
    \centering
    \includegraphics[width=0.96\linewidth]{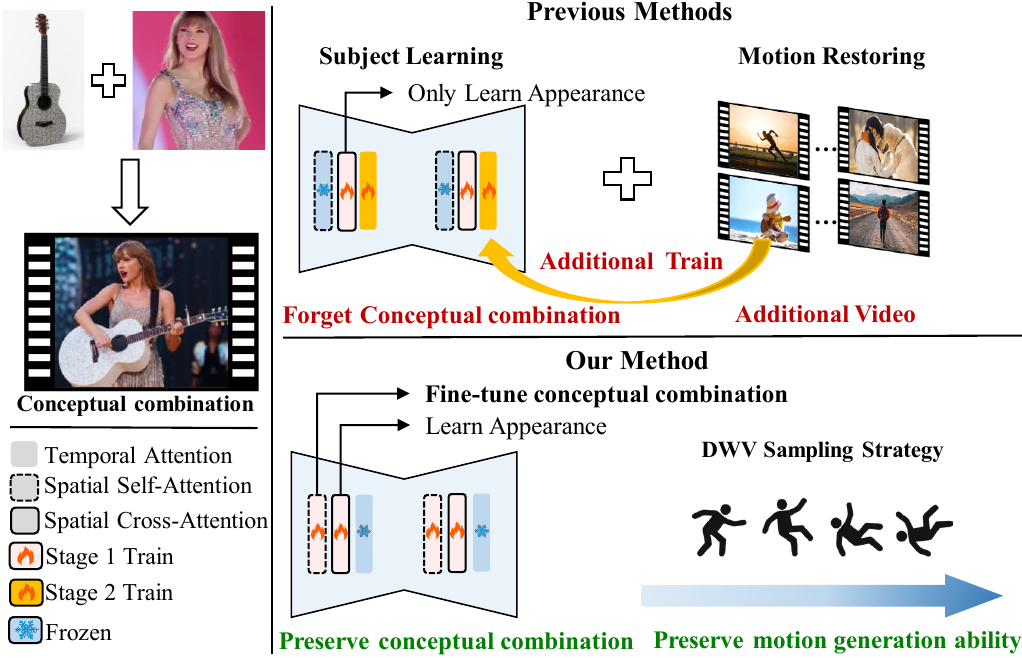}
    \caption{Comparison of our approach with previous work. Our method can better learn the appearance  of the subject while preserving  the concept combination ability and motion generation ability, only requires one stage of training without additional videos. DWV Sampling Strategy is our Dynamic Weighted Video Sampling Strategy.}
    \label{fig:intro}
\end{figure}

\begin{figure*}[h!]
  \centering
  \includegraphics[width=0.96\linewidth]{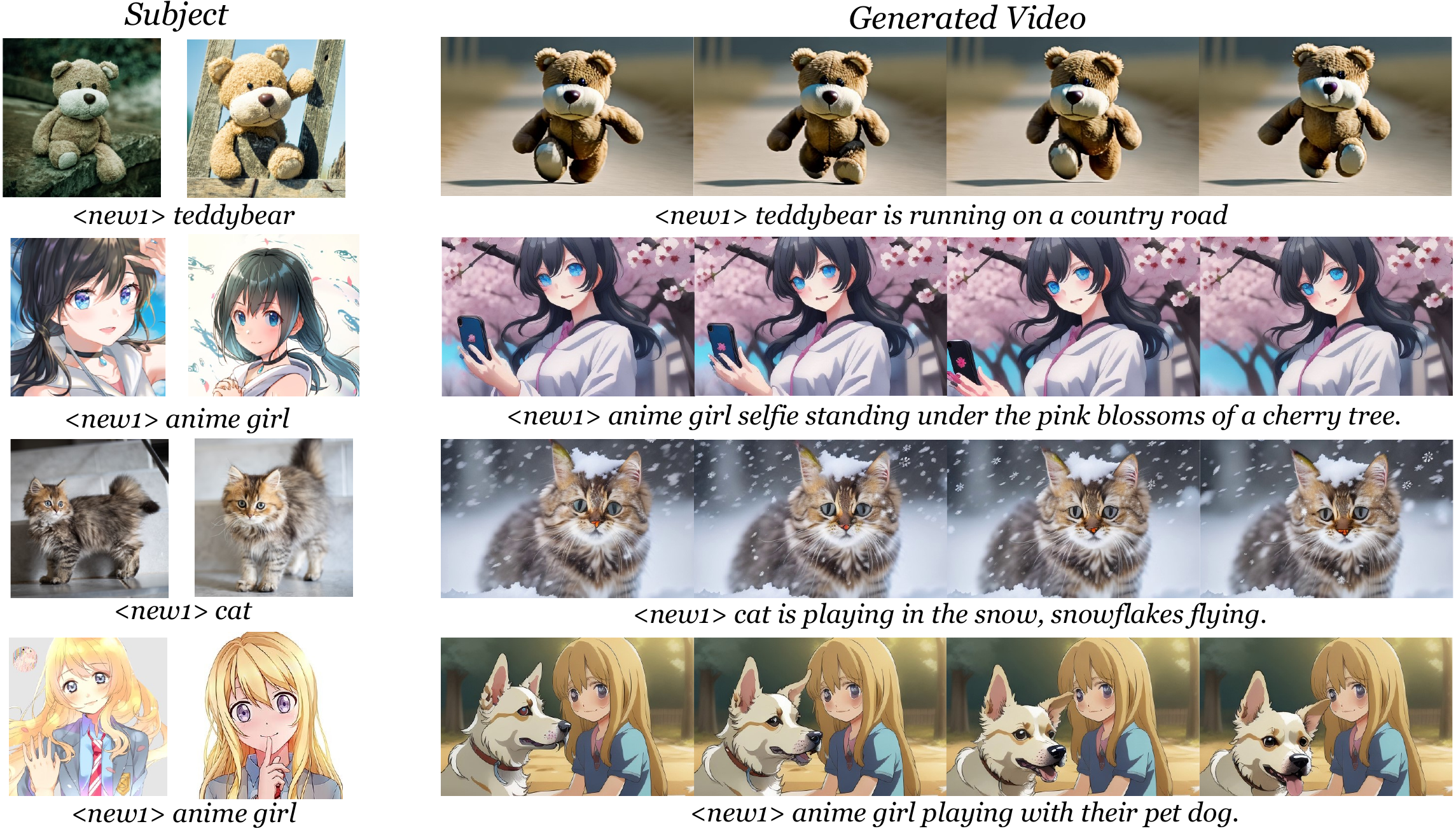}
  \caption{Visualization for our CustomCrafter. Our approach allows customization of subject identity and movement patterns to generate the desired video with text prompt by preserving motion generation and conceptual combination abilities.}
  \label{fig:first_fig}
\end{figure*}
With the development of diffusion models and multimodal, text-to-video generation has made significant progress~\cite{videoworldsimulators2024,yang2023bridging,yang2024solving,yang2024rcs,miao2023self,su2023referring,su2023language}.
Challenges still arise when users want to generate videos of specific subjects.  
Customized video generation needs to simultaneously satisfy three requirements: consistent subject appearance, free concept combination, and smooth motion generation. 
Concept combination refers to the ability to combine the learned specific subject with other different concepts.
For example, as shown in ~\Cref{fig:intro}, when we learn about a specific guitar, we hope that this guitar can be combined with other concepts (e.g., a person) to generate videos.
Numerous studies~\cite{gal2022image,ruiz2023dreambooth,han2023svdiff,gu2024mix,kumari2023multi,ruiz2023hyperdreambooth,huang2024context,huang2024group} have proposed many methods for customized image generation and have achieved good results.
When these approaches are applied directly to customized video generation, they often fail to generate videos well.
These methods damage the conceptual combination ability and motion generation ability of the text-to-video model during the fine-tuning process, which means that the subject learned by the model cannot be combined with other concepts and the motion in the generated video tends to be static.

Recent methods for customized video generation have noticed these issues. 
Some works~\cite{wei2023dreamvideo,he2023animate} have recognized the decline in motion fluidity and conceptual combination abilities. 
These methods believe that the damage to the model's concept combination ability and motion generation ability cannot be recovered, which is caused by using images to fine-tune the VDMs during subject learning.
So, as shown in ~\Cref{fig:intro}, they use further fine-tuning of the model parameters with additional videos similar to the content described in the prompt~\cite{wei2023dreamvideo}, or use videos to guide the video generation process~\cite{he2023animate}.
However, these methods require retrieving similar text prompts from massive video libraries to generate different prompts for the same subject. 
It is often necessary to frequently change the guiding videos or even re-fine-tune the model, leading to additional training, which brings great inconvenience to users.
Considering the above issues, one question naturally arises: \textbf{Is it possible to generate videos of a specified subject only by performing subject learning, while preserving the model's inherent abilities of concept combination and motion generation?}

In this paper, we introduce a novel framework, CustomCrafter, which preserves the model's motion generation and conceptual combination abilities without the need for additional video and fine-tuning to recover these abilities. 
Through experiments, we have observed that:
(1) Numerous studies~\cite{cao2023masactrl,gu2024photoswap,nam2024dreammatcher} have indicated that for image models, self-attention often significantly affects the ability to combine concepts. 
And self-attention plays a crucial role in preserving geometric and shape for subject~\cite{liu2024towards}.
We have observed that this phenomenon remains applicable to VDMs.
Furthermore, influenced by ~\cite{kumari2023multi}, existing work only updates spatial cross-attention during subject learning.
(2) During the video diffusion models (VDMs) generation process, we can observe that in different timesteps of denoising process, the model repair content has a certain tendency. 
The early stages of the denoising process tend to restore the layout of each frame and the motion, whereas the later stages focus on the recovery of object detail.

Therefore, to address the issue of decreased concept combination ability, we propose the Spatial Subject Learning Module, which can update the weights of both the spatial cross-attention and self-attention layers during fine-tuning.
This improves the model's ability to capture the appearance of new subjects, while also improving the model's ability to combine new subjects with other concepts.
Regarding the decline in motion generation ability, we design the Spatial Subject Learning Module to be pluggable, and propose the Dynamic Weighted Video Sampling Strategy, which improves the model's inference process.
When generating videos, by leveraging the pluggable nature of the subject learning module, we can preserve the model's motion generation ability by reducing the influence of subject learning on the stages that tend to restore motion. 
Then, during the stages that tend to repair details, we can restore the influence of the subject learning module. 
This ensures the consistency of the subject's appearance, thereby enabling the generation of videos of a specified subject using the model's inherent motion generation ability.
As shown in ~\Cref{fig:first_fig}, we can generate high-quality videos of a specified subject by preserving the inherent abilities of VDMs of concept combination and motion generation and only performing subject learning.
Our method does not require the introduction of additional videos as guidance or repeated fine-tuning of the model. 
By preserving the inherent knowledge of the text-to-video model, we can conveniently generate videos of specified objects that align with the prompt.

Through extensive experiments, we provide qualitative, quantitative, and user study results that demonstrate the superiority of our method in customized video generation. 
Our contributions are summarized as follows:
\begin{itemize}

    \item As far as we know, we are the first to discover and use the property of VDMs' denoising process to decouple appearance and motion to improve customized generation.
    \item We propose a subject learning method that can learn the appearance of the subject better and effectively preserve the ability to combine new subjects with other concepts.
    \item We introduce a sampling strategy that can preserve the motion generation of VDMs without using additional videos to guide or fine-tune the model.

\end{itemize}

\section{Related Work}
\subsection{Text-to-Video Diffusion Models}
Diffusion models~\cite{sohl2015deep,ho2020denoising,jiang2024survey} have recently emerged as a trend in generative models, particularly in the domain of text-to-image (T2I) generation~\cite{ramesh2022hierarchical,rombach2022high,zhang2023real,wu2024spherediffusion,wang2024target,wu2024ifadapter,huang2024group,zheng2023layoutdiffusion}. 
For video generation, Video Diffusion Model has been introduced to model video distributions.
Pioneering work to utilize a space-time-factored U-Net for video modeling in pixel space for unconditional video generation was done by VDM~\cite{ho2022video,dou2024gvdiff}.
AnimateDiff~\cite{guo2023animatediff} further advanced the field of text-to-video generation by incorporating a motion module into the Stable Diffusion model. 
Following this, LVDM~\cite{he2022latent,chen2023videocrafter1} suggest extending LDM to model videos in the latent space of an auto-encoder.
This method gradually became mainstream and derived many methods, including ModelScope~\cite{wang2023modelscope}, LAVIE~\cite{wang2023lavie}, PYOCO~\cite{ge2023preserve}, VideoFactory~\cite{wang2023videofactory}, VPDM~\cite{yu2023video}, and VideoCrafter2~\cite{chen2024videocrafter2}.
%

\subsection{Customized Generation on Diffusion Models}
Numerous studies~\cite{wei2023elite,li2024blip} have proposed many methods for custom image generation and achieved good results. 
Most current work focuses on subject customization with a few images~\cite{ han2023svdiff, gu2024mix, shi2023instantbooth}.
Moreover, some works study the more challenging multi-subject customization task~\cite{yuan2023inserting, xiao2023fastcomposer, ma2023subject, chen2023anydoor}.
Despite significant progress in customized image generation, customized video generation is still under exploration.
Although there have been initial attempts to customize video diffusion, such as VideoAssembler ~\cite{zhao2023videoassembler}, VideoBooth ~\cite{jiang2023videobooth},ID-Animator~\cite{he2024id} and CustomVideo~\cite{wang2024customvideo}, which use reference images to personalize the video diffusion model while preserving the identity of the subject.
These approaches focus on addressing the problem of generating videos with a similar subject appearance, neglecting disruption to motion and conceptual combination abilities.
DreamVideo ~\cite{wei2023dreamvideo} first decouples the learning process for subject and motion. 
Animate-A-Story~\cite{he2023animate} refers to the depth information of the additional video to guide motion generation.
However, the use of additional video data and the need to retrain the model according to different prompts bring great inconvenience to users.

\section{Preliminary}
\textbf{Video diffusion models (VDMs)}~\cite{wang2023modelscope,he2022latent, guo2023animatediff, chen2024videocrafter2}2 are designed for video generation tasks by extending image diffusion models to adapt to video data. VDMs learn a video data distribution by the gradual denoising of a variable sampled from a Gaussian distribution. 
First, a learnable autoencoder (consisting of an encoder $\textbf{E}$ and a decoder $\textbf{D}$) is trained to compress the video into a smaller latent space representation.
Then, a latent representation $z = \textbf{E}(x)$ is trained instead of a video $x$.
Specifically, the diffusion model $\epsilon_{\theta}$ aims to predict the added noise $\epsilon$ at each timestep $t$ based on the text condition $c$, where $t \in \mathcal{U}(0, 1)$. The training objective can be simplified as a reconstruction loss:
\begin{equation}
    \mathcal{L}_{video} = \mathbb{E}_{z, c, \epsilon \sim \mathcal{N}(0, \mathbf{\mathrm{I}}), t}\left[\left\| \epsilon - \epsilon_\theta\left(z_t, c, t\right)\right\|_2^2\right],
\label{eq:diffusion_loss}
\end{equation}
where $z \in \mathbb{R}^{F \times H \times W \times C}$ is the latent code of video data with $F, H, W, C$ being frame, height, width, and channel, respectively. $\tau_\theta$ presents a pre-trained text encoder.
$c$ is the text prompt for input video.
A noise-corrupted latent code $z_t$ from the ground-truth $z_0$ is formulated as $z_t = \lambda_t z_0 + \sigma_t \epsilon$, where $\sigma_t = \sqrt{1 - \lambda_t^2}$, $\lambda_t$ and $\sigma_t$ are hyperparameters to control the diffusion process. 
In this work, we have selected the VideoCrafter2~\cite{chen2024videocrafter2} as our base model.
\section{Method}

\begin{figure}[tb]
    \centering
    \includegraphics[width=0.96\linewidth]{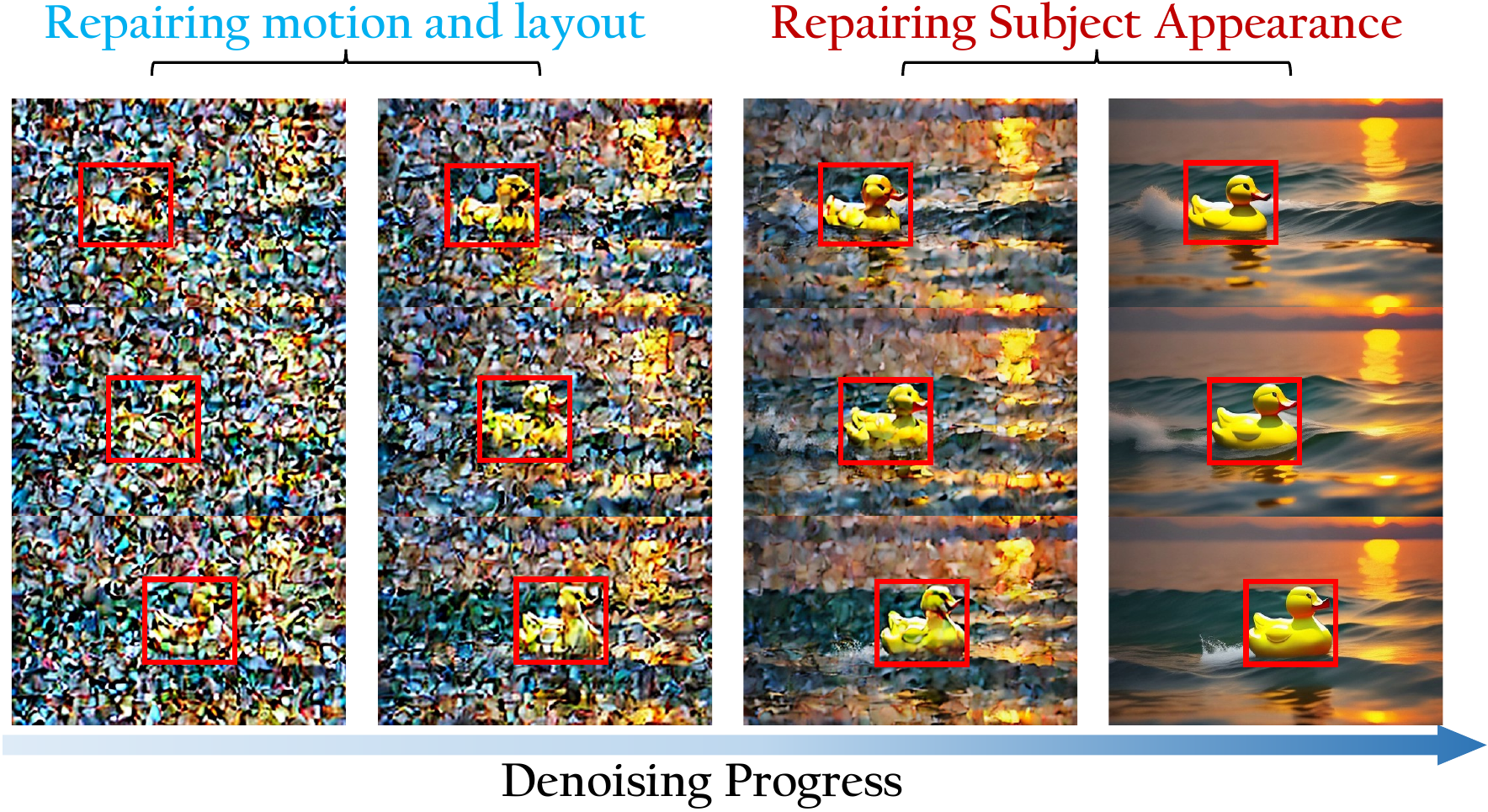}
    \caption{Visualization of video denoising process. The motion is formed in early stages of the denoising process, and the subject's appearance emerges in later stages.
}
    \label{fig:vis_denoising}
\end{figure}
\begin{figure*}[tb]
    \centering
    \includegraphics[width=0.98\linewidth]{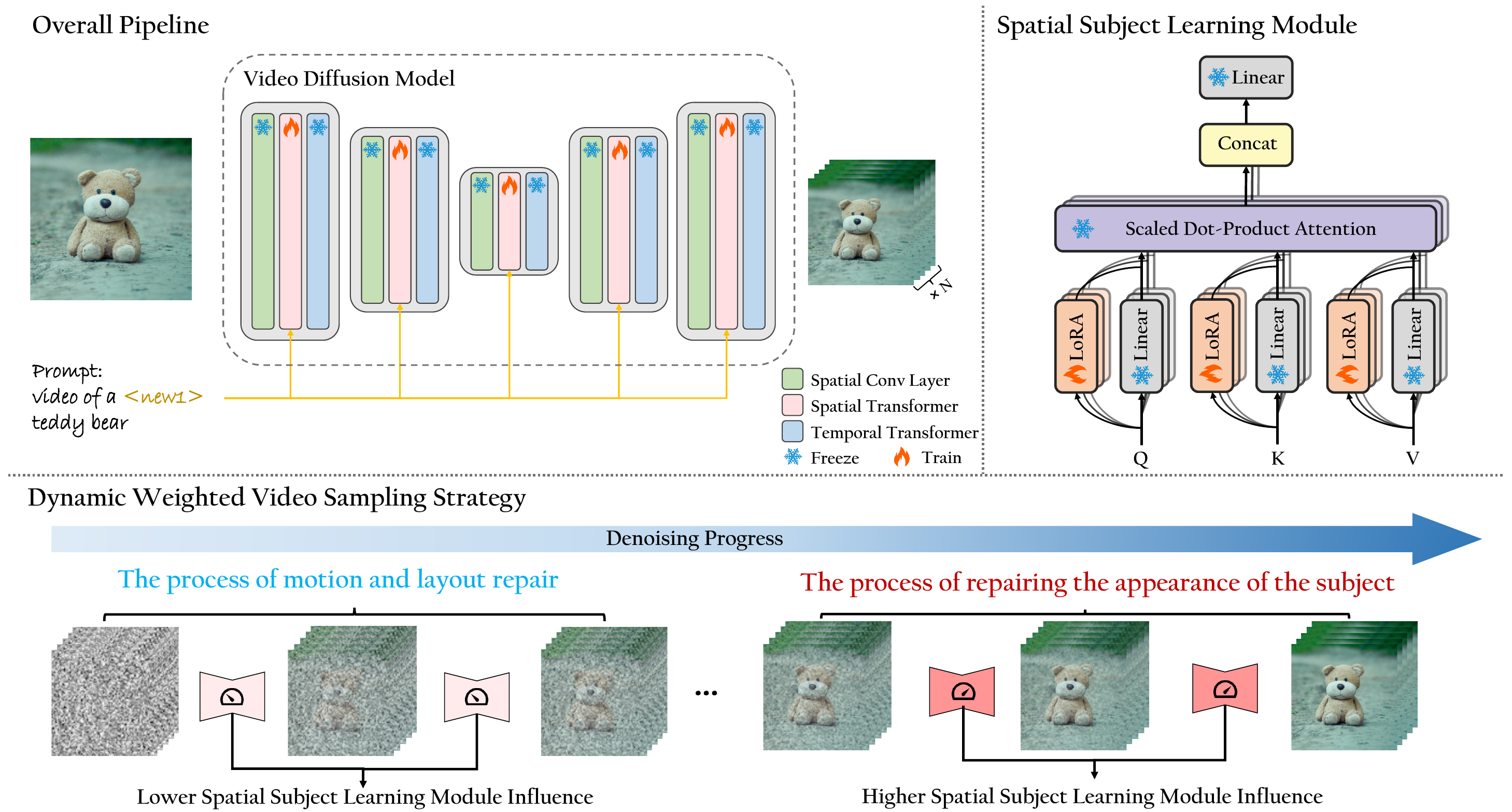}
    \caption{Overall review of CustomCrafter. For subject learning, we adopt LoRA to construct Spatial Subject Learning Module, which update the Query, Key, and Value parameters of attention layers in all Spatial Transformer models. In the process of generating videos, we divide the denoising process into two phases: the motion layout repair process and the subject appearance repair process. By reducing the influence of the Spatial Subject Learning Module in the motion layout repair process, and restoring it in the subject appearance repair process to repair the details of the subject.
}
    \label{fig:pipline}
\end{figure*}

In this section, we first analyze the reasons for the impairment of conceptual combination and motion capabilities in models during the subject learning and outline our findings about the VDMs.
Then, we introduce the Spatial Subject Learning Module, which used to learn the appearance details of the subject. 
Finally, we introduce our Dynamic Weighted Video Sampling Strategy, which can generate high-quality videos without additional video guidance or training after the model has learned the appearance of the subject.

\subsection{Explore the Video Diffusion Model}

Given a pre-trained VDM, our goal is to enable the model to learn the subject's appearance from a number of images and the corresponding text prompt. 
To achieve this goal, most existing works update part of the model to learn subject's appearance. 
However, since the training data only consist of images and the VDMs require a relatively long training period to learn a new concept, the model inevitably forgets its motion generation and concept combination ability.

Regarding the issue of conceptual combination, many studies on diffusion models point out ~\cite{cao2023masactrl,nam2024dreammatcher} that for text-to-image diffusion models, self-attention significantly affects the geometric and shape for subject and the ability to combine concepts. 
We found the same phenomenon in video diffusion models. 
However, current methods tend to fine-tune only the parameters of cross-attention during training. 
This results in the model being unable to learn the appearance of new subjects and reduces its ability to combine new subjects with other concepts.

For motion generation, as shown in ~\Cref{fig:vis_denoising}, we observed that the duck's movement is formed early in the denoising process. 
The later stage of the denoising process is to enhance the appearance of the subject based on the established motion. 
Therefore, during the video generation process, the models repair content with a certain tendency. 
The VDMs tend to restore the overall layout and motion in the early stages of the denoising process, while focusing on the recovery of object detail in the later stages. 
Therefore, our approach is to utilize a plug-and-play module to facilitate subject learning. 
By reducing the influence of this module on the denoising generation process in the early stages of inference, we can mitigate the disruption to the motion generation capability of VDMs. 
In the later stages of the denoising process, where object detail recovery occurs, we increase the influence of this module. 
This approach allows us to preserve the original model's video motion generation ability while generating high-quality appearance of new subjects.

\subsection{Spatial Subject Learning Module}
For learning the appearance of new subject, we constructed a Spatial Subject Learning Module.
During training, we repeat a single picture of the object $N$ times to turn the picture into a still $N$ frame video.
As shown in ~\Cref{fig:pipline}, our fine-tuning parameters can be divided into two parts. 
First, following textual inversion~\cite{gal2022image}, we employ a new token $V^{*}$ and learn a new token embedding vector in the CLIP text encoder to represent a new concept. 
For example, a specified teddy bear can be represented as $V^{*} teddybear$.
The second part pertains to the spatial transformer module of the video diffusion model. 
We fine-tuned both the cross-attention module and the self-attention module in the spatial transformer blocks of VDM to ensure that the model has the ability to combine new subjects with other concepts while learning the appearance of the subject.
To achieve a plug-and-play effect, we adopted the Low-Rank Adaptation~\cite{hu2021lora} (LoRA) method for fine-tuning.
LoRA applies a residue path of two low-rank matrices $B\in\mathbb{R}^{d\times r}, A\in\mathbb{R}^{r\times k}$ in the attention layers, whose original weight is $W_{0}\in\mathbb{R}^{d\times k}$, $r\ll\min(d, k)$.
The new forward path is as follows:
\begin{equation}
    W = W_{0} + \lambda \Delta W = W_{0} + \lambda B A,
\label{eq:LoRA}
\end{equation}
where $\lambda$ is a coefficient adjusting the strength of the added LoRA.
In this paper, we insert LoRA layers into the query, key, and value corresponding to $W_{q}$,$W_{k}$,$W_{v}$ in both the cross-attention and self-attention modules for fine-tuning.

\subsection{Dynamic Weighted Video Sampling Strategy}

To alleviate the decrease in motion generation ability, as shown in \Cref{al:dwvss}, we propose Dynamic Weighted Video Sampling Strategy.
We notice that during the generation process of the video diffusion model, in the early stage of denoising, the model tends to restore the video's motions. 
In the later stage of the denoising process, the model tends to restore the details of the generated video. 
Therefore, in the first $K$ steps of the denoising process, we adjust the weight $\lambda$ of all LoRA modules in our Spatial Subject Learning Module to a smaller value $\lambda_{s}$, to ensure that the model is almost unaffected by motion stagnation and the decrease in the combination ability of concepts caused by the parameters of the subject's overfit. 
In the later stage of the denoising process, we restore the weight $\lambda$ of the LoRA modules to a higher value $\lambda_{l}$, allowing the model to further repair the specific details of each frame of the subject, thereby generating high-quality videos of the specified subject.
\begin{algorithm}
    \textbf{Input:} A source prompt $\mathcal{P}$, a random seed $s$, a small LoRA weight $\lambda_s$ used in Phase 1, a large LoRA weight $\lambda_l$ used in Phase 2, and the delimitation point $k$.
    
    \textbf{Output:} latent code for generating video.
    
     $z_{T} \sim N(0,I)$ a unit Gaussian random variable with random seed $s$;
     
     $Change(DM, \lambda, \lambda_{s})$;  \tcc{Change $\lambda$ to $\lambda_{s}$}
     
     \For{$t=T,T-1,\ldots,1$}
     {
        \If{t == ( T - K )}{
          $Change(DM, \lambda, \lambda_{l})$ \tcc{Change $\lambda$  to $\lambda_{l}$}
        }
        $z_{t-1} \gets DM(z_{t},\mathcal{P},t,s)$\;
     }
     \textbf{Return:} $z_{0}$
     \caption{Dynamic Weighted Video Sampling Strategy}
     \label{al:dwvss}
\end{algorithm}

\subsection{Model Training Strategy}
Inspired by previous work~\cite{ruiz2023dreambooth,kumari2023multi}, during training, we use class-specific prior preservation to mitigate overfitting issues in the training process, to enhance the diversity of the generated videos.
The loss for prior preservation is formulated as the following:
\begin{align}
    \mathcal{L}_{pr} = \mathbb{E}_{z^{pr}, c^{pr}, \epsilon \sim \mathcal{N}(0, \mathbf{\mathrm{I}}), t}\left[\left\| \epsilon - \epsilon_\theta\left(z_t^{pr}, c^{pr}, t\right)\right\|_2^2\right],
\label{eq:pr_diffusion_loss}
\end{align}
where $z^{pr}$ is the latent code of the input regularized video, $c^{pr}$ is the text condition for the input regularized video.
In training, our total loss function is as follows:
\begin{equation}
    \mathcal{L} = \mathcal{L}_{video}+\alpha\mathcal{L}_{pr},
\end{equation}
where $\alpha$ is a hyper-parameter to adjust the relative weight of prior-preservation.
\section{Experiments}

\subsection{Experimental Setup}
\subsubsection{Datasets and Protocols}
For subject customization, we select subjects from image customization papers~\cite{ruiz2023dreambooth,kumari2023multi} for a total of 20 subjects. 
For each subject, we use ChatGPT to generate 10 related prompts, which are used to test the generation of specified motion videos for the subject.
All experiments use VideoCrafter2 as the base model. 
When learning the subject, we use the AdamW optimizer, set the learning rate to $3 \times 10^{-5}$ and the weight decay to $1 \times 10^{-2}$. 
We perform 10,000 iterations on 4 NVIDIA A100 GPUs.
For the Class-specific Prior Preservation Loss, similar to~\cite{ruiz2023dreambooth,kumari2023multi},we collected 200 images from LAION-400M~\cite{schuhmann2021laion} for each subject as regularization data and set $\alpha$ to $1.0$.
During the inference process, we use DDIM~\cite{song2020denoising} for 50-step sampling and classifier-free guidance with a cfg of 12.0 to generate videos with a resolution of $512 \times 320$.
For all subjects, to facilitate experimentation and comparison, we uniformly set $\lambda_{s}$ and $\lambda_{l}$ to 0.4 and 0.8 respectively, and set $K$ to 5 based on our observation. In actual use, these parameters can be adjusted by the user.
\begin{table}[]
    \centering
    \resizebox{0.48\textwidth}{!}{
    \begin{tabular}{lcccc}
    \toprule
    Method            & CLIP-T$ \uparrow$         & CLIP-I $\uparrow$         & DINO-I$ \uparrow$         & T. Cons. $\uparrow$       \\ \midrule
    Custom Diffusion  & 0.289          & 0.759          & 0.546          & 0.990          \\
    Custom Diffusion* & 0.286          & 0.769          & 0.583          & 0.992          \\
    DreamVideo        & 0.298          & 0.724          & 0.489          & 0.992          \\
    DreamVideo*       & 0.295          & 0.748          & 0.536          & 0.993          \\
    \textbf{Ours}     & \textbf{0.318} & \textbf{0.786} & \textbf{0.627} & \textbf{0.994}   \\ \bottomrule
    \end{tabular}
    }
    \caption{Comparison with the existing methods. Note that Custom Diffusion* and DreamVideo* in the table represent the results we get after extending the number of training steps in the original paper.}
    \label{tab:sota}
\end{table}
\begin{figure*}[tb]
    \centering
    \includegraphics[width=1.0\linewidth]{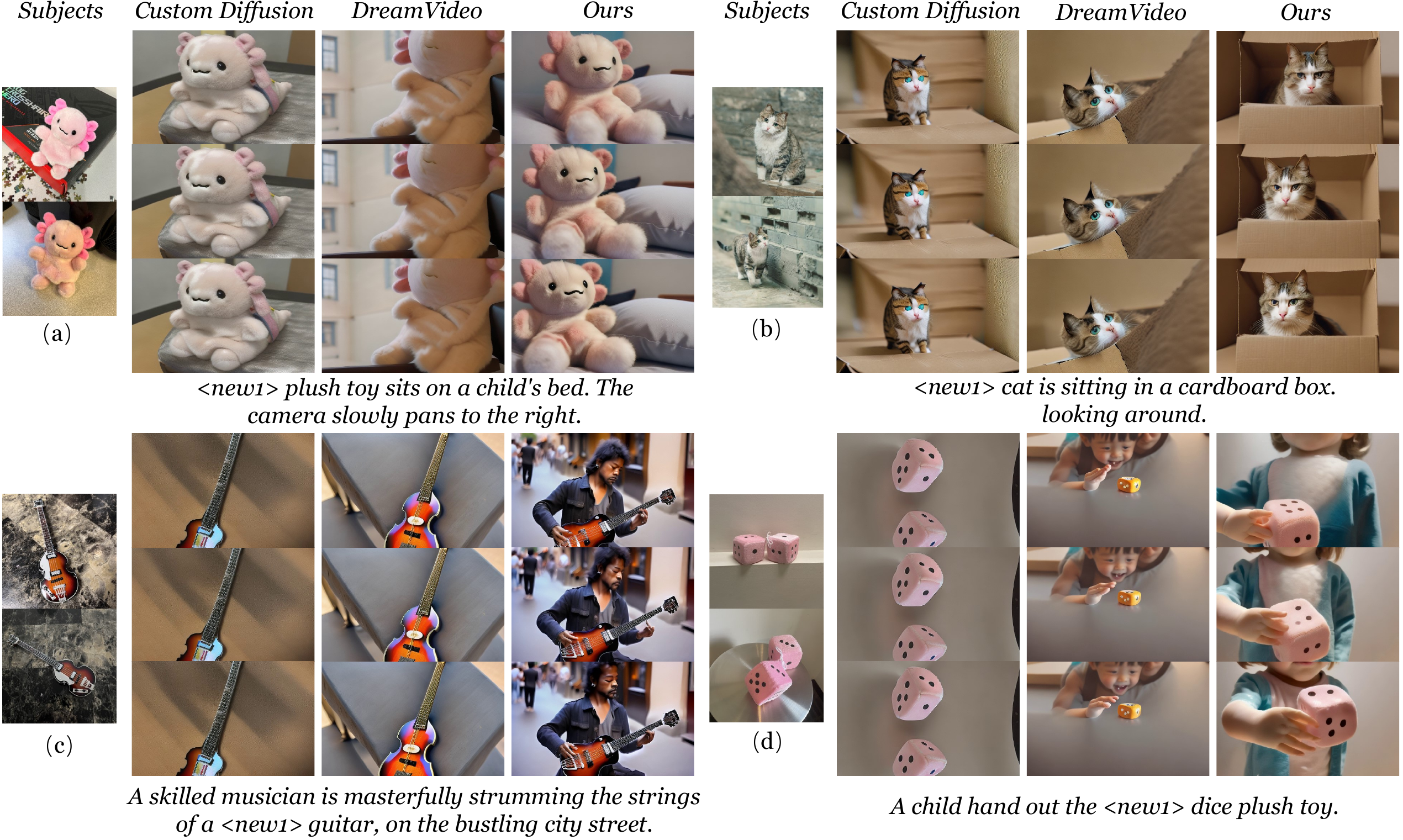}
    \caption{Qualitative comparison of customized video generation with both subjects and motions. Without guidance from additional videos, our method significantly outperforms in terms of concept combination.}
    \label{fig:main_cmp}
\end{figure*}

\subsubsection{Baselines}
Given that different base model are chosen in the current field of video customization, we reproduce Custom Diffusion~\cite{kumari2023multi} and DreamVideo~\cite{wei2023dreamvideo} based on VideoCrafter2. 
Since our methods do not introduce additional videos as guidance, to ensure fairness, we only reproduce the subject learning part of DreamVideo for fair comparison. 
In addition, considering that VDMs need more steps to learn the appearance of the subject, and the default settings of Custom Diffusion and DreamVideo cannot fit the subject appearance features well, we accordingly extend the training steps of these methods.

\subsubsection{Evaluation Metrics}
Follow~\cite{wei2023dreamvideo,wang2024customvideo}, we evaluate our approach with the following four metrics:
(1) \textit{CLIP-T} calculates the average cosine similarity between CLIP~\cite{radford2021learning} image embeddings of all generated frames and their text embedding.
(2) \textit{CLIP-I} measures the visual similarity between the generated and target subjects. We computed the average cosine similarity between the CLIP image embeddings of all generated frames and the target images.
(3) \textit{DINO-I}~\cite{ruiz2023dreambooth}, another metric to measure visual similarity using ViTS/16 DINO~\cite{zhang2022dino}. Compared to CLIP, the self-supervised training model encourages the distinguishing features of individual subjects.
(4) \textit{Temporal Consistency}~\cite{esser2023structure}, we compute CLIP image embeddings on all generated frames and report the average cosine similarity between all pairs of consecutive frames.

\subsection{Quantitative Results}

We trained 20 subjects using Custom Diffusion, DreamVideo and our method, respectively. 
After training, we used each method to generate videos for each subject using 10 prompts, employing the same random seed and denoising steps. 
The results, as shown in ~\Cref{tab:sota}, indicate that our method outperforms existing methods in all four metrics. 
The degree of text alignment and subject fidelity have been significantly improved. 
The temporal consistency of the generated videos is roughly equivalent to that of other methods. 
The metrics used to evaluate the subject fidelity, CLIP-I and DINO-I, have improved by 1.7\% and 4.4\%, respectively, compared to existing methods. 
The degree of text alignment has improved by 1.5\% compared to the previous best result.

\subsection{Qualitative Results}
We also visualized some results for qualitative analysis. 
We used the prompt of dynamic videos to generate videos of specified subjects, observing the subject fidelity in the generated videos and the motion fluency.  
As shown in ~\Cref{fig:main_cmp}(a), when we want to generate a video of a specified plush toy sitting on a child's bed and the camera slowly pans to the right, we find that existing methods overfit reference image during training.
Without guidance from additional videos, the generated motions are almost static. 
However, our method can generate videos with fluent motions and right concept combination.
Besides, in ~\Cref{fig:main_cmp}(b), only our method correctly generates the conceptual combination of the cat and the cardboard box and the motion of ``looking around" with high subject fidelity. 
Furthermore, in ~\Cref{fig:main_cmp}(c), when we want to generate a video of a musician playing a given guitar, we find that existing methods greatly damage the model's ability to combine concepts. 
They can only generate the specified subject, but cannot generate a musician playing the guitar, and the motion is "frozen". 
Similarly, in ~\Cref{fig:main_cmp}(d), when we want to generate a video of a child handing out the dice toy, a similar situation occurs.
Our method successfully generated the combination of concept of a child and a toy dice, and has smooth motions.
Therefore, without guidance from additional videos, our method significantly outperforms existing methods in terms of concept combination ability and motion naturalness, and has better subject fidelity.
Please refer to the supplementary material for more visualizations and demonstration videos.

\subsection{User study}
To further validate the effectiveness of our method, we conducted a human evaluation of our method and existing methods without using additional video data as guidance. 
\begin{figure}[tb]
    \centering
    \includegraphics[width=0.98\linewidth]{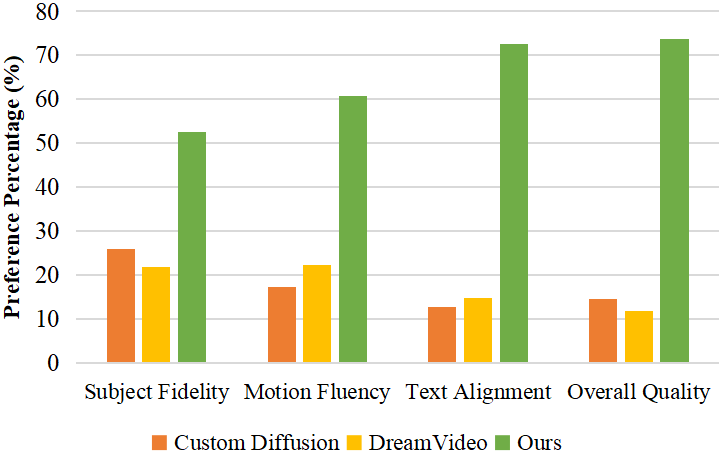}
    \caption{User Study. Our CustomCrafter achieves the best human preference compared with other comparison methods}
    \label{fig:user}
\end{figure}
We invited 20 professionals to evaluate the 30 sets of generated video results.
For each group, we provided subject images and videos generated using the same seed and the same text prompt under different methods for comparison. 
We evaluated the quality of the generated videos in four dimensions: Text Alignment, Subject Fidelity, Motion Fluency, and Overall Quality.
Text Alignment evaluates whether the generated video matches the text prompt. 
Subject Fidelity measures whether the generated object is close to the reference image. 
Motion Fluency is used to evaluate the quality of the motions in the generated video. 
Overall Quality is used to measure whether the quality of the generated video overall meets user expectations.
As shown in ~\Cref{fig:user}, our method has gained significantly more user preference in all metrics, proving the effectiveness of our method.

\subsection{Ablation Study}
\begin{figure}[tb]
    \centering
    \includegraphics[width=0.98\linewidth]{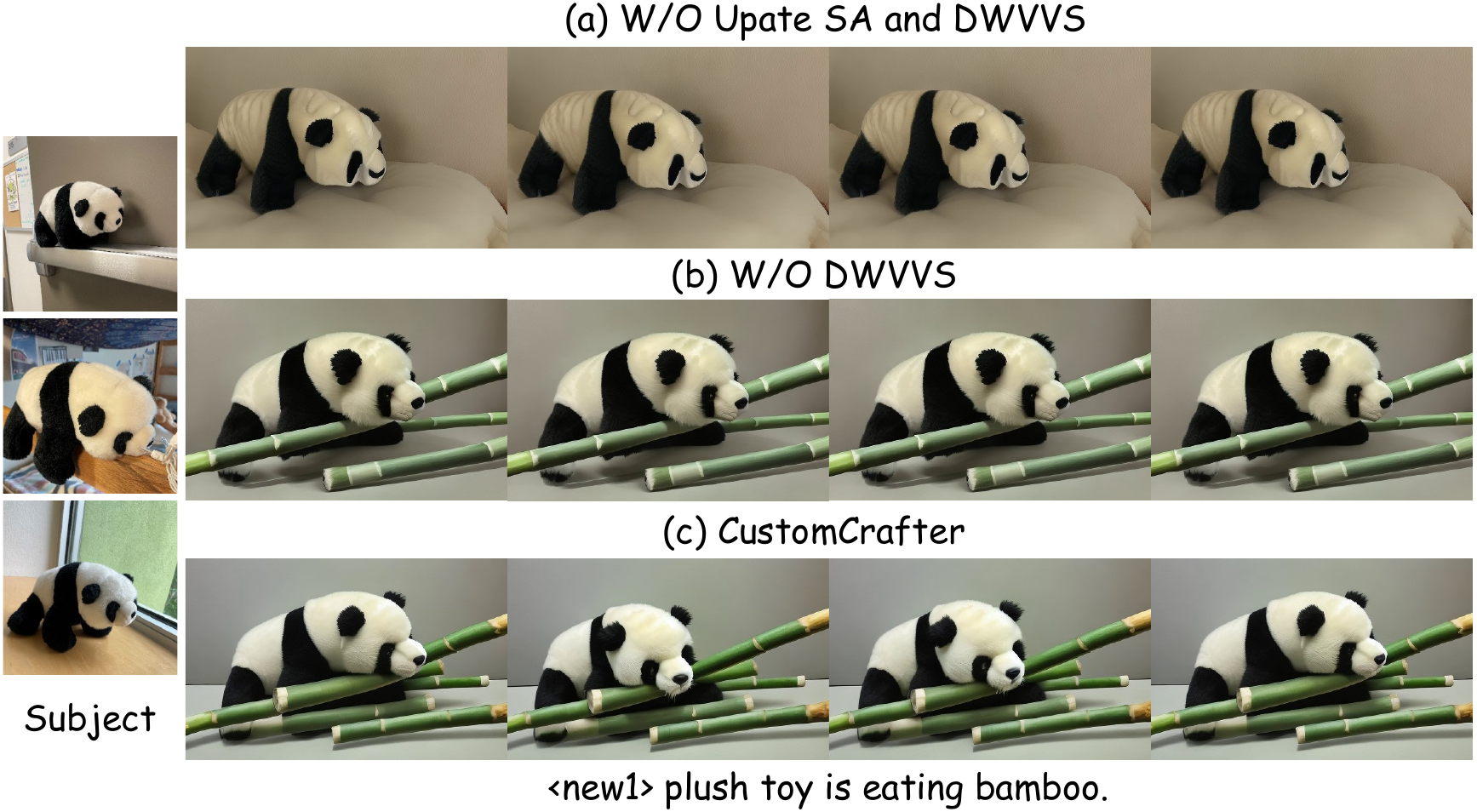}
    \caption{Effect of each design of our method. It can be seen that each of our designs has achieved the expected effect.}
    \label{fig:ablation}
\end{figure}

\begin{table}[]
    \small
    \centering
    \resizebox{0.48\textwidth}{!}{
    \begin{tabular}{cc|cccc}
    \toprule
    SSLM       & DWVSS      & CLIP-T$\uparrow$ & CLIP-I$\uparrow$ & DINO-I$\uparrow$ & T.Cons.$\uparrow$ \\ \midrule
               &            & 0.286  & 0.769  & 0.583  & 0.992   \\
    \checkmark &            & 0.294  & 0.790  & 0.631  & 0.993   \\
    \checkmark & \checkmark & 0.318  & 0.786  & 0.627  & 0.994   \\ \bottomrule
    \end{tabular}}
    \captionsetup{font={small}}
    \caption{Ablation Study. ``SSLM" is Spatial Subject Learning Module, ``DWVSS" is Dynamic Weighted Video Sampling Strategy.}
    \label{tab:abalation2}
\end{table}
In this section, we construct ablation studies to validate the specific roles and effectiveness of each component that our proposed.
As shown in ~\Cref{tab:abalation2}, we choose Custom Diffusion as the baseline to present the quantitative results of our designed Spatial Subject Learning Module and Dynamic Weighted Video Sampling Strategy, respectively. 
It can be observed that using our Spatial Subject Learning Module achieves better results on the CLIP-I and DINO-I metrics, which measure the subject fidelity. 
This suggests that compared to previous work, our method is more capable of capturing the details of the subject. 
The Dynamic Weighted Video Sampling Strategy, due to modifications in the process, may result in a slight impairment of the subject's appearance similarity. 
However, the motions can be significantly improved, substantially enhancing the text alignment.
In addition, we use visualization results to demonstrate the effectiveness of our method.
As shown in ~\Cref{fig:ablation}, when we only update the parameters of cross-attention, we find that the model's ability to combine concepts is poor and it cannot generate a simple concept combination of pandas and bamboo that matches the prompt. 
However, when we use our Spatial Subject Learning Module module to update both cross-attention and self-attention, and do not use the Dynamic Weighted Video Sampling Strategy, the generated video's ability to combine concepts is improved, but the VDM cannot generate fluent motions that match the text prompt.
After adopting our sampling strategy, the generated video has almost no significant loss in subject fidelity, but the naturalness of the motion has greatly improved.

\section{Conclusion}
In this paper, we introduce our CustomCrafter, a novel framework for customized video generation. 
This approach does not require additional video to repair motion generation ability.
We first designed a Spatial Subject Learning Module, which updates the Spatial Attention for subject learning. 
Simultaneously, we proposed a Dynamic Weighted Video Sampling Strategy, which improves the model's inference process to restore the motion generation capability of VDMs.
Through experiments, we have demonstrated that our method is better than existing methods.

\section*{Acknowledgements}
This work is supported in part by National Science Foundation for Distinguished Young Scholars under Grant 62225605, Project 12326608 supported by NSFC, Natural Science Foundation of Shanghai under Grant 24ZR1425600, Zhejiang Provincial Natural Science Foundation of China under Grant LD24F020016, ``Pioneer" and ``Leading Goose" R\&D Program of Zhejiang (No. 2024C01020), as well as the Ningbo Science and Technology Innovation Project (No.2024Z294), and sponsored by CCF-Tencent Rhino-Bird Open Research Fund and Research Fund of ARC Lab, Tencent PCG.
\bibliography{aaai25}
\end{document}